\def\year{2020}\relax
\documentclass[letterpaper]{article} 
\pdfoutput=1

\usepackage{aaai20}  
\usepackage{times}  
\usepackage{helvet}  
\usepackage{courier}  
\usepackage{url}  
\usepackage{graphicx}  
\usepackage[normalem]{ulem}
\usepackage[utf8]{inputenc} 
\usepackage[T1]{fontenc}    
\usepackage{booktabs}       
\usepackage{nicefrac}       
\usepackage{microtype}      
\usepackage{verbatim}
\usepackage{bbm}            
\usepackage{xcolor}
\usepackage{amsfonts}       
\usepackage{amsmath}
\usepackage{amssymb}
\usepackage{mathtools}

\DeclareMathOperator*{\argmax}{arg\,max}
\DeclarePairedDelimiterX{\infdivx}[2]{(}{)}{%
  #1\;\delimsize\|\;#2%
}

\DeclarePairedDelimiter{\ceil}{\lceil}{\rceil}

\newcommand{\pres}[1]{#1^\text{\normalfont pres}}

\newcommand{\disc}[2]{\bar{#1}_{(#2)}}

\newcommand{\prop}[2]{\tilde{#1}_{(#2)}}

\newcommand{\select}[2]{#1_{(#2)}}

\newcommand{\sni}[3]{#1^\text{\normalfont #2}_{#3}}
\newcommand{\sn}[2]{#1^\text{\normalfont #2}}
\newcommand{\snm}[2]{#1^#2}

\newcommand{\citet}[1]{\citeauthor{#1}~\shortcite{#1}}
\newcommand{\citep}{\cite}

\graphicspath{{.}}

\title{Exploiting Spatial Invariance for Scalable Unsupervised Object Tracking}
\author{
  Eric Crawford\\
  Mila/McGill University\\
  Montreal, QC, Canada\\
  \texttt{eric.crawford@mail.mcgill.ca} 
  \And
  Joelle Pineau\\
  Mila/McGill University\\
  Montreal, QC, Canada\\
  \texttt{jpineau@cs.mcgill.ca} \\
}

\begin{document}
\maketitle
\begin{abstract}
    The ability to detect and track objects in the visual world is a crucial skill for any intelligent agent, as it is a necessary precursor to any object-level reasoning process. Moreover, it is important that agents learn to track objects without supervision (i.e. without access to annotated training videos) since this will allow agents to begin operating in new environments with minimal human assistance. The task of learning to discover and track objects in videos, which we call \textit{unsupervised object tracking}, has grown in prominence in recent years; however, most architectures that address it still struggle to deal with large scenes containing many objects. In the current work, we propose an architecture that scales well to the large-scene, many-object setting by employing spatially invariant computations (convolutions and spatial attention) and representations (a spatially local object specification scheme). In a series of experiments, we demonstrate a number of attractive features of our architecture; most notably, that it outperforms competing methods at tracking objects in cluttered scenes with many objects, and that it can generalize well to videos that are larger and/or contain more objects than videos encountered during training.
\end{abstract}

\section{Introduction}
The ability to reason about objects is a vital skill for intelligent agents. Indeed, it underlies much of human intelligence, and is one of several ``core'' domains of human cognition, meaning that it is sufficiently important that evolution has endowed humans (and many animals) with special-purpose hardware for discovering, tracking and reasoning about objects \cite{carey2009origin}. Recently, significant progress has been made in the design of neural networks that can reason about objects, and these architectures have been shown to possess a number of significant advantages over systems that lack object-like representations \citep{diuk2008object,chang2016compositional,kansky2017schema,zambaldi2018relational}. However, in order to reason in terms of objects, a system first needs a way of discovering and tracking objects in the world; moreover, the system should be able to do this without supervision, in order to be able to adapt to new environments with little human oversight. Much of the recent work on object-level reasoning assumes that object-like representations are directly provided by the environment, thereby avoiding the problem of discovering and learning to track objects.

In the current work, we aim to make progress on this task of discovering and tracking objects, which we call \textit{unsupervised object tracking}. Our general approach is to formulate a Variational Autoencoder (VAE) \citep{kingma2013auto} for videos, endowed with a highly structured, object-like latent representation. This VAE is composed of a number of modules that are applied each timestep of the input video, notably a discovery module which detects new objects in the current frame, and a propagation module which updates the attributes of objects discovered in previous frames based on information from the current frame. For each input frame, a rendering module creates a corresponding output frame from the objects proposed by the discovery and propagation modules. The network is trained by maximizing the VAE evidence lower bound, which encourages the output frames to be accurate reconstructions of the input frames. At the end of training, it is expected that the discovery module will have become a competent object detector, while the propagation module will have learned object dynamics. This overall architecture was first proposed in Sequential Attend Infer Repeat (SQAIR) \citep{kosiorek2018sequential}.

However, as we demonstrate empirically, SQAIR struggles at processing spatially large videos that contain many densely packed objects. We hypothesize that this is because SQAIR does not fully exploit the spatial statistics of objects. One example is SQAIR's discovery module, which processes input frames holistically. Specifically, it computes an initial representation for an input frame by processing it with a Multi-layer Perceptron (MLP), effectively discarding all spatial structure, before recurrently emitting a sequence of detected objects. The discovery module is consequently unable to exploit the fact that objects are \textit{spatially local}. A consequence of this spatial locality is that in order to detect an object, a local detector can be used which only has access to a sub-region of the frame. Doing so has the notable advantage that a sub-region will have significantly less complexity than the frame as a whole, making the detector's job easier. In order to ensure the entire frame is covered, the local detector can be applied repeatedly to different (but possibly overlapping) sub-regions. This scheme can be implemented efficiently through the use of convolutional neural networks; indeed, this is the intuition behind the success of recent supervised single-shot object detectors such as SSD \citep{liu2016ssd} and YOLO \citep{redmon2017yolo9000}.

In the current work, we propose \textbf{S}patially \textbf{I}nvariant \textbf{L}abel-free \textbf{O}bject \textbf{T}racking (SILOT) (pronounced like ``silo''), a differentiable architecture for unsupervised object tracking that is able to scale well to large scenes containing many objects. SILOT achieves this scalability by making extensive use of spatially invariant computations and representations, thereby fully exploiting the structure of objects in images. For example, in its discovery module SILOT employs a convolutional object detector with a spatially local object specification scheme, and spatial attention is used throughout to compute objects features in a spatially invariant manner.

Through a number of experiments, we demonstrate the concrete advantages that arise from this focus on spatial invariance. In particular, we show that SILOT has a greatly improved capacity for handling large, many-object videos, and that trained SILOT networks can generalize well to videos that are larger and/or contain different numbers of objects than videos encountered in training.

\section{Related Work}
Our work builds on a growing body of recent research on learning to detect and track objects without supervision. The pioneering work Attend, Infer, Repeat (AIR) \citep{eslami2016attend} formulated a VAE for images, with a highly structured, object-like latent representation. By training this VAE to reconstruct input images, the encoder module of the VAE is forced to learn to detect objects. AIR has since been extended in a number of directions. Sequential Attend, Infer, Repeat (SQAIR) \citep{kosiorek2018sequential} extended AIR to handle videos rather than images. This was achieved by combining an object discovery module, which extracts objects from the scene and has a design similar to AIR, with an object propagation module which tracks changes in object attributes (e.g. position, size, appearance) over time.
Similar extensions are also provided by \citep{hsieh2018learning,he2018tracking}. In an orthogonal direction, Spatially Invariant Attend, Infer, Repeat (SPAIR) \citep{crawford2019spatially} improved on AIR's ability to handle cluttered scenes by replacing AIR's recurrent encoder network with a convolutional network and a spatially local object specification scheme. Finally, Discrete-AIR \citep{wang2019unsupervised} showed how to adapt AIR to discover discrete object classes.

AIR and its descendants generally describe each object using, at minimum, a bounding box and the appearance of the object within that box. A separate but related body of work uses an alternate scheme, specifying each object using two image-sized maps: a mask map, which determines how much the object ``owns'' each image pixel, and an appearance map, which specifies the object's appearance at pixels that it owns. In these architectures, objects compete with one another for ownership of image pixels through their mask maps.
Examples include \citep{greff2017neural,van2018relational,goel2018unsupervised,greff2019multi,burgess2019monet}.
One weakness of this style of object specification is that the objects are relatively heavy-weight, which makes it computationally intensive to use large numbers of objects. Additionally, there is no way to ensure that the objects are compact and object-like (in AIR-style architectures, a prior is used to encourage discovered objects to have small spatial extent). Together these weaknesses make these architectures poorly suited to handling large scenes with many objects. For instance, when the architecture proposed in \citep{goel2018unsupervised} was applied to scenes from the Space Invaders Atari game, it grouped the 36 aliens into a single object, rather than allocating a separate object per alien.

As a final note, an architecture similar to SILOT, with a similar emphasis on ability to scale to large numbers of objects, was concurrently developed in \cite{jiang2019scalable}.

\section{Spatially Invariant, Label-free Object Tracking}
    SILOT is a Variational Auotencoder (VAE) which models a video as a collection of moving objects. The model is divided into modules: a discovery module, which detects objects from each frame; a propagation module, which updates the attributes of previously discovered objects; a selection module, which selects a small set of objects to keep from the union of the discovered and propagated objects; and a rendering module which renders selected objects into an output frame. The discovery, propagation and selection modules constitute the VAE encoder $h_\phi(z | x)$, with all encoder parameters gathered together in a vector $\phi$. Meanwhile the rendering module constitutes the VAE decoder $g_\theta(x | z)$, with its parameters collected in a vector $\theta$.

    \subsection{Overview}
    In this section we give a high-level overview of the architecture of SILOT. Throughout this work we generally surround temporal indices with brackets to differentiate them from other types of indices (e.g. indexing a specific object within a set of objects). To reduce clutter we often drop the temporal index when it can be inferred from the context.
    
    Assume we are given a length-$T$ input video $\{x_{(t)}\}_{t=0}^{T-1}$, with $x_{(t)}$ denoting an individual frame for $t \in \{0, \dots, T-1\}$. For each timestep, we consider a number of variable sets:
    \begin{enumerate}
        \item Discovered latents $\disc{z}{t}$ and discovered objects $\disc{o}{t}$
        \item Propagated latents $\prop{z}{t}$ and propagated objects $\prop{o}{t}$ 
        \item Selected objects $\select{o}{t}$
    \end{enumerate}
    The total set of latent variables for the VAE is:
    \begin{align*}
    z = \bigcup_{t=0}^{T-1} \disc{z}{t} \cup \prop{z}{t}
    \end{align*}
    For both discovery and propagation, the object sets are deterministic functions of the corresponding latent variable sets (and possibly other variables). The choice to employ a separate set of latents, apart from the objects, was made in part because it is generally inconvenient to define prior distributions (required for training the VAE) in the highly structured, interpretable space of objects. Past architectures have employed a similar approach \citep{eslami2016attend,kosiorek2018sequential}.

\begin{figure}[!ht]
        \centering
        \includegraphics[width=\columnwidth]{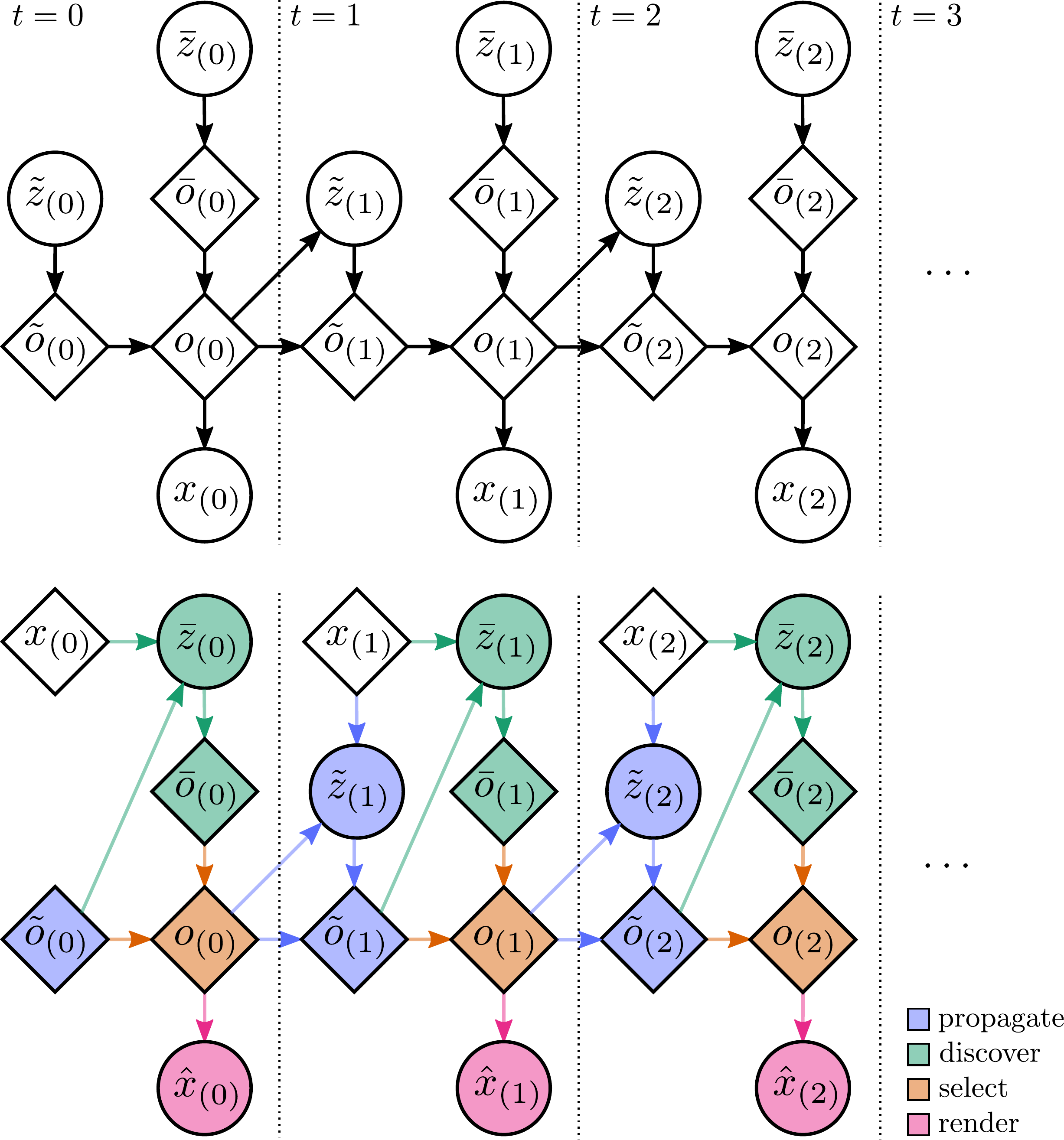}
        \caption{High-level overview of SILOT. Top: Generative model assumed by SILOT. Diamonds/circles are deterministic/stochastic functions of their inputs. Bottom: Structure of the SILOT neural network. Modules are indicated by color. Propagation, discovery and selection constitute the encoder/inference network $h_\phi(z | x)$, while rendering constitutes the decoder/likelihood network $g_\theta(x | z)$. \label{fig:high-level}}
    \end{figure}

    The generative model assumed by SILOT, as well as the high-level structure of the SILOT neural network, is shown in Figure \ref{fig:high-level}. There we see the relationships between the variables and modules discussed previously. Both the generative model and high-level network structure are similar to and inspired by SQAIR \citep{kosiorek2018sequential}. Within a timestep $t$, the flow of computation proceeds as follows:
    \begin{enumerate}
        \item Input frame $x_{(t)}$ and objects from the previous frame $\select{o}{t-1}$ are passed into the propagation module, which predicts a set of updates $\prop{z}{t}$ and deterministically applies them, yielding propagated objects $\prop{o}{t}$.
        \item $x_{(t)}$ and $\prop{o}{t}$ are passed into the discovery module, which discovers objects in the frame that are not accounted for by any propagated object, first yielding $\disc{z}{t}$ and then $\disc{o}{t}$ via a deterministic transformation.
        \item $\prop{o}{t}$ and $\disc{o}{t}$ are passed into the selection module, which selects a subset of the objects to retain, yielding $\select{o}{t}$.
        \item $\select{o}{t}$ is passed into the rendering module which yields an output frame $\hat{x}_{(t)}$.
    \end{enumerate}
    On the initial timestep, the propagation module is not executed and a degenerate set of objects is used for $\tilde{o}_{(0)}$.

    \subsection{Object Representation}\label{sec:object}
        In SILOT, an individual object is represented by a collection of variables, each called an \textit{attribute}:
        \begin{align*}
            \sn{o}{where}& \in \mathbb{R}^4& \sn{o}{what}& \in \mathbb{R}^A& \sn{o}{depth}& \in [0, 1] & \sn{o}{pres}& \in [0, 1]
        \end{align*}
        $\sn{o}{where}$ decomposes as $\sn{o}{where} = (\snm{o}{y}, \snm{o}{x}, \snm{o}{h}, \snm{o}{w})$. $\snm{o}{y}$ and $\snm{o}{x}$ specify the location of the object's center, while $\snm{o}{h}$ and $\snm{o}{w}$ specify the object's size.
        $\sn{o}{what}$ acts as a catch-all, storing information about the object that is not captured by other attributes (e.g. appearance, velocity).
        $\sn{o}{depth}$ specifies the relative depth of the object; in the output image, objects with higher values for this attribute appear on top of objects with lower values.
        $\sn{o}{pres}$ specifies the extent to which the object exists; objects with $\sn{o}{pres} = 0$ do not appear in the output image.
    \subsection{Discovery}\label{sec:discovery}
        The role of the object discovery module is to take in the current frame $x_{(t)}$ and the set of objects propagated from the previous frame $\tilde{o}_{(t-1)}$, and detect any objects in the frame that are not yet accounted for
        (either because those objects appeared for the first time in the current frame, or because they were not previously discovered due to error).
        Object discovery in SILOT is implemented as a convolutional neural network in order to achieve spatial invariance, and is heavily inspired by SPAIR, an unsupervised convolutional object detector \citep{crawford2019spatially}, as well as by single-shot supervised object detectors such as SSD and YOLO.

\begin{figure}
            \centering
            \includegraphics[width=0.9\columnwidth]{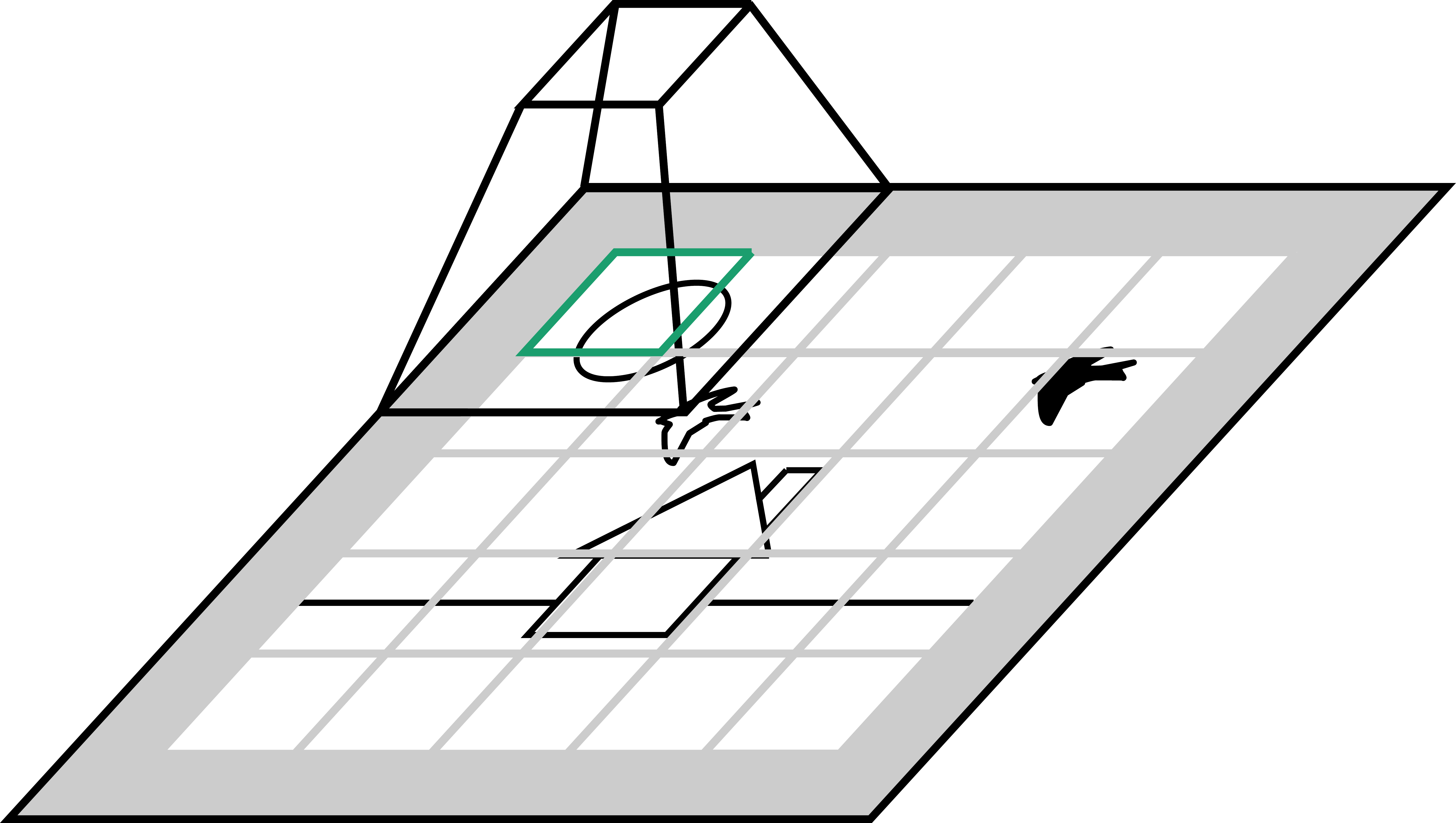}
            \caption{
                Schematic depicting grid cells and padding in the discovery module. The gray grid is the grid of cells described in Section 3.3. The top of the trapezoid is a spatial location in the output layer of $\sni{d}{bu}{\phi}$, associated with the grid cell highlighted in green. The bottom of the trapezoid is the receptive field of that spatial location; the frame is padded (solid gray border area) before being passed into $\sni{d}{bu}{\phi}$ to ensure that all receptive fields are centered on their corresponding grid cells.
                \label{fig:grid}
            }
        \end{figure}

        An initial convolutional network $\sni{d}{bu}{\phi}$ extracts ``bottom-up'' information from the current input frame $x_{(t)}$, mapping to a feature volume $\sni{v}{bu}{(t)}$:
        \begin{align*}
            \sni{v}{bu}{(t)} &= \sni{d}{bu}{\phi}(x_{(t)})
        \end{align*}
        The structure of network $\sni{d}{bu}{\phi}$ can be taken to induce a spatial grid over the frame, as follows. Let $c_h/c_w$ be the translation (in pixels) vertically/horizontally between receptive fields of adjacent spatial locations in $\sni{v}{bu}{(t)}$.
        Then for an input frame with dimensions $(H_\text{inp}, W_\text{inp}, 3)$, we divide the frame up into an $(H, W)$ grid of cells, each cell being $c_h$ pixels high by $c_w$ pixels wide, where $H = \ceil{H_\text{inp} / c_h}$, $W = \ceil{W_\text{inp} / c_w}$. The output volume $\sni{v}{bu}{(t)}$ has spatial shape $(H, W)$, and we associate each of its spatial locations with the corresponding grid cell. Importantly, the input frame is padded on all sides to ensure that the receptive field for each spatial location in $\sni{v}{bu}{t}$ is centered on the corresponding grid cell\footnote{\citep{dangha2017guide} is a useful guide to receptive field arithmetic in convolutional nets.}. This scheme is visualized in Figure \ref{fig:grid}.

        \begin{figure}[!ht]
            \centering
            \includegraphics[width=0.85\columnwidth]{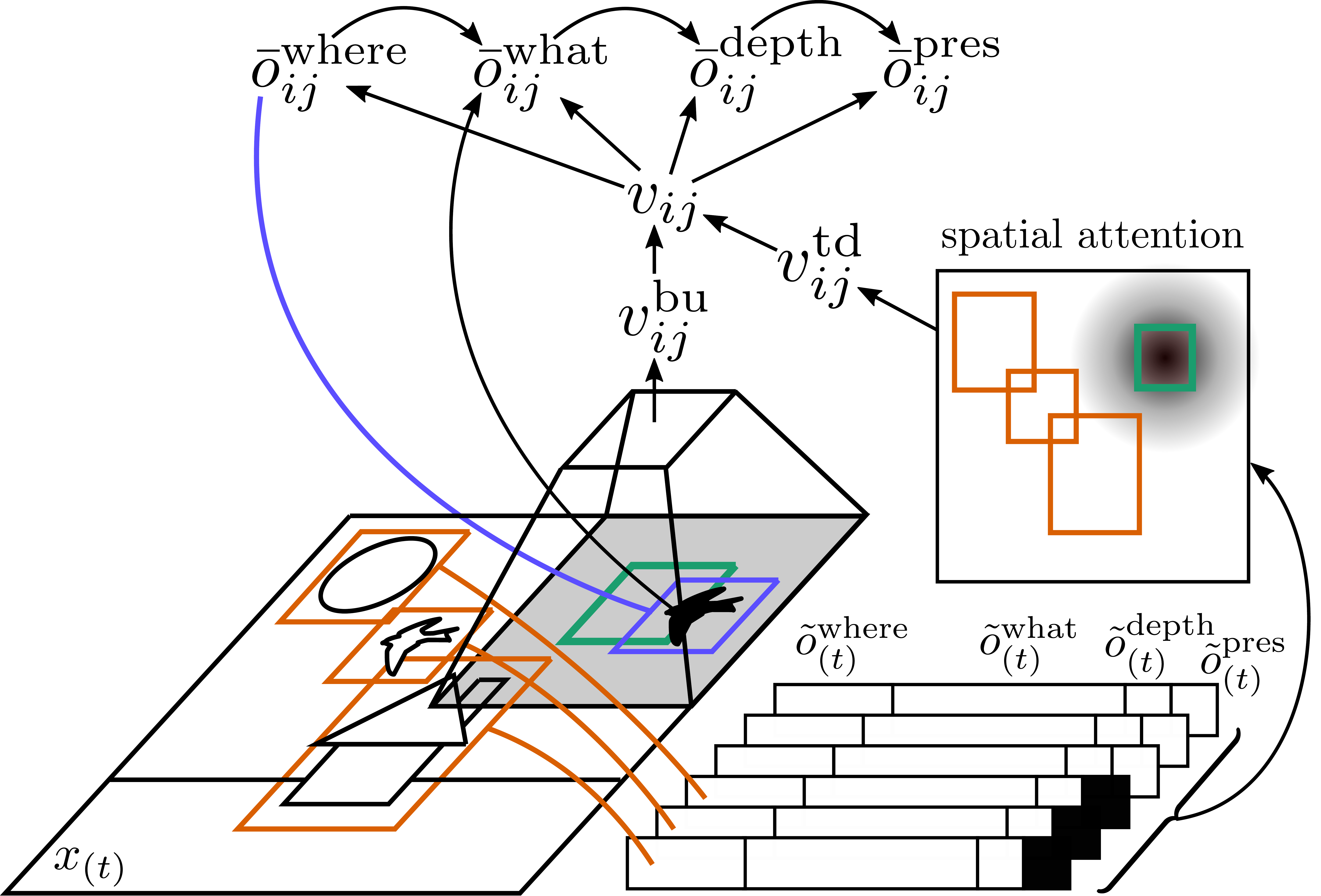}
            \caption{Schematic depicting the structure of a discovery unit with indices $ij$ at time $t$, discovering the black bird which has just come into view from the right. Local bottom-up information from the current frame is processed by a convolutional filter (trapezoid), which has a receptive field (grey base of the trapezoid) centered on the discovery unit's grid cell (green rectangle). Next, top-down information about nearby objects propagated from the previous frame (orange boxes) is summarized using spatial attention with a Gaussian kernel centered at the grid cell. Bottom-up and top-down information is then fused and used to autoregressively predict object attributes (here we have omitted the latent discovery variables $\bar{z}_{ij}$). $\sni{\bar{o}}{where}{ij}$ is specified with respect to the grid cell, and the prediction for $\bar{o}^\text{what}_{ij}$ conditions on the output of a spatial transformer parameterized by $\sni{\bar{o}}{where}{ij}$ (blue rectangle).
            \label{fig:discovery}
            }
        \end{figure}

        The discovery module will ultimately yield a separate object for each grid cell. It is useful to think of the discovery module as consisting of a 2D array of identical object detectors, each operating on a different local region of the image. We call each of these local detectors a \textit{discovery unit}; the structure of a unit is shown in Figure \ref{fig:discovery}.
        
        Variables for all units are grouped together into convolutional volumes, and computations are implemented by size-preserving convolutions on these volumes, essentially computing all discovery units in parallel. Thus, unless otherwise specified, all variables in this section are volumes with spatial dimensions $(H, W)$, and all networks are convolutional and preserve spatial dimensions. The convolutional layers use a kernel width and stride of 1 so that each discovery unit is independent (such a layer is equivalent to applying a fully connected layer separately to each spatial location).
        
        In order to avoid rediscovering objects that are already accounted for, each discovery unit needs to be aware of ``top-down'' information about objects propagated from the previous frame that are near its grid cell. We thus employ a spatial attention step which, for each grid cell, extracts features of propagated objects, and then weights those features according to a Gaussian kernel centered at the cell (similar to the Neural Physics Engine \citep{chang2016compositional}). The result is a feature volume $\sni{v}{td}{(t)}$ with spatial shape $(H, W)$ where each spatial location contains information about nearby propagated objects:

        \begin{align*}
            \sni{v}{td}{(t)} &= \text{SpatialAttention}^\text{disc}_\phi(\tilde{o}_{(t)}, \sigma)
        \end{align*}
        where $\sigma$ is a hyperparameter giving the standard deviation of the Gaussian kernel. Details on the spatial attention step are provided in Section B of the Appendix.

        Next, a convolutional network $\sni{d}{fuse}{\phi}$ combines bottom-up and top-down information (here we begin omitting temporal indices):
        \begin{align*}
            v &= \sni{d}{fuse}{\phi}(\sn{v}{bu}, \sn{v}{td})
        \end{align*}
        We follow the convention that a neural network that takes multiple inputs first concatenates those inputs along their trailing dimension (i.e. the depth dimension in the case of convolutional volumes), and processes the result as usual.

        The network then predicts parameters for distributions over the latent discovery variables $\bar{z}$, samples from the predicted distributions, and then maps the sampled latent to the more interpretable $\bar{o}$. This is done on an attribute-by-attribute basis (in order $[$where, what, depth, pres$]$), 
        and is autoregressive, so that predictions for later attributes are conditioned on samples for earlier attributes. These steps are detailed below.

        \subsubsection{Predicting $\sn{\bar{o}}{where}$\normalfont{.}}
            We first use a network $\sni{d}{where}{\phi}$ to predict parameters for a distribution over $\sn{\bar{z}}{where}$, and then sample:
            \begin{align*}
                \sn{\bar{\mu}}{where}, \sn{\bar{\sigma}}{where} &= \sni{d}{where}{\phi}(v)\\
                \sn{\bar{z}}{where} &\sim N(\sn{\bar{\mu}}{where}, \sn{\bar{\sigma}}{where})
            \end{align*}
            Next we deterministically map to $\sn{\bar{o}}{where}$. We decompose $\sn{\bar{z}}{where}$ as $\sn{\bar{z}}{where} = (\bar{z}^{y}, \bar{z}^{x}, \bar{z}^{h}, \bar{z}^{w})$. Here it will be useful to narrow our focus to a single discovery unit with indices $ij$ for $i \in \{0, \dots, H-1\}$, $j \in \{0, \dots, W-1\}$. 
            $\bar{z}^{y}_{ij}$ and $\bar{z}^{x}_{ij}$ parameterize the position of the object according to:
            \begin{align*}
                {b}^{y}_{ij} &= b^\text{min} + \text{sigmoid}(\bar{z}^{y}_{ij}) \left( b^\text{max} - b^\text{min}\right)\\
                \bar{o}^{y}_{ij} &= (i + b^{y}_{ij}) c_h\\
                b^{x}_{ij} &= b^\text{min} + \text{sigmoid}(\bar{z}^{x}_{ij}) \left( b^\text{max} - b^\text{min}\right)\\
                \bar{o}^{x}_{ij} &= (j + b^{x}_{ij}) c_w
            \end{align*}
            where $b^\text{min}$ and $b^\text{max}$ are fixed real numbers which, in effect, impose bounds on the distance between the object and the grid cell.
            $\bar{z}^{h}_{ij}$ and $\bar{z}^{w}_{ij}$ parameterize the size of the object as:
            \begin{align*}
                \bar{o}^{h}_{ij} &= \text{sigmoid}(\bar{z}^{h}_{ij}) a_h & \bar{o}^{w}_{ij} &= \text{sigmoid}(\bar{z}^{w}_{ij}) a_w
            \end{align*}
            for fixed real numbers $a_h$ and $a_w$. $(a_h, a_w)$ can be interpreted as the dimensions of an \textit{anchor box} as used in supervised object detection \citep{ren2015faster}. Specifying object size with respect to $(a_h, a_w)$, as opposed to the size of the input frame, ensures that $\bar{o}^{h}_{ij}$ and $\bar{o}^{w}_{ij}$ are meaningful regardless of the spatial dimensions of the input frame.

            \subsubsection{Predicting $\sn{\bar{o}}{what}$, $\sn{\bar{o}}{depth}$ and $\sn{\bar{o}}{pres}$\normalfont{.}}
                In order to obtain highly location-specific information from the image, an array of glimpses $\bar{g}$ (one glimpse per discovery unit) is extracted from the image using spatial transformers $\tau$ \citep{jaderberg2015spatial} parameterized by $\sn{\bar{o}}{where}$. These glimpses are then mapped to a feature volume $v^\text{obj}$ by a network $\sni{d}{obj}{\phi}$:
                \begin{align*}
                    \bar{g} &= \tau(x, \sn{\bar{o}}{where})\\
                    \sn{v}{obj} &= \sni{d}{obj}{\phi}(\bar{g})
                \end{align*}
                We then autoregressively predict the remaining attributes:
                \begin{align*}
                    \sn{\bar{\mu}}{what}, \sn{\bar{\sigma}}{what} &= \sni{d}{what}{\phi}(v, \sn{v}{obj}, \sn{\bar{o}}{where})&\\
                    \sn{\bar{z}}{what} &\sim N(\sn{\bar{\mu}}{what}, \sn{\bar{\sigma}}{what})&\\
                    \sn{\bar{o}}{what} &= \sn{\bar{z}}{what}\\
                    \sn{\bar{\mu}}{depth}, \sn{\bar{\sigma}}{depth} &= \sni{d}{depth}{\phi}(v, \sn{v}{obj}, \sn{\bar{o}}{where}, \sn{\bar{o}}{what})\\
                    \sn{\bar{z}}{depth} &\sim N(\sn{\bar{\mu}}{depth}, \sn{\bar{\sigma}}{depth})\\
                    \sn{\bar{o}}{depth} &= \text{sigmoid}(\sn{\bar{z}}{depth})\\
                    \sn{\bar{\mu}}{pres} &= \sni{d}{pres}{\phi}(v, \sn{v}{obj}, \sn{\bar{o}}{where}, \sn{\bar{o}}{what}, \sn{\bar{o}}{depth})\\
                    \sn{\bar{z}}{pres} &\sim \text{Logistic}(\sn{\bar{\mu}}{pres})\\
                    \sn{\bar{o}}{pres} &= \text{sigmoid}(\sn{\bar{z}}{pres})
                \end{align*}
                Note that $\sn{\bar{o}}{pres}$ can be viewed as a set of \text{BinConcrete} random variables \citep{maddison2016concrete}. The significance of this choice is discussed below in Section \ref{sec:select}.

        \subsection{Propagation}
\begin{figure}
                \centering
                \includegraphics[width=\columnwidth]{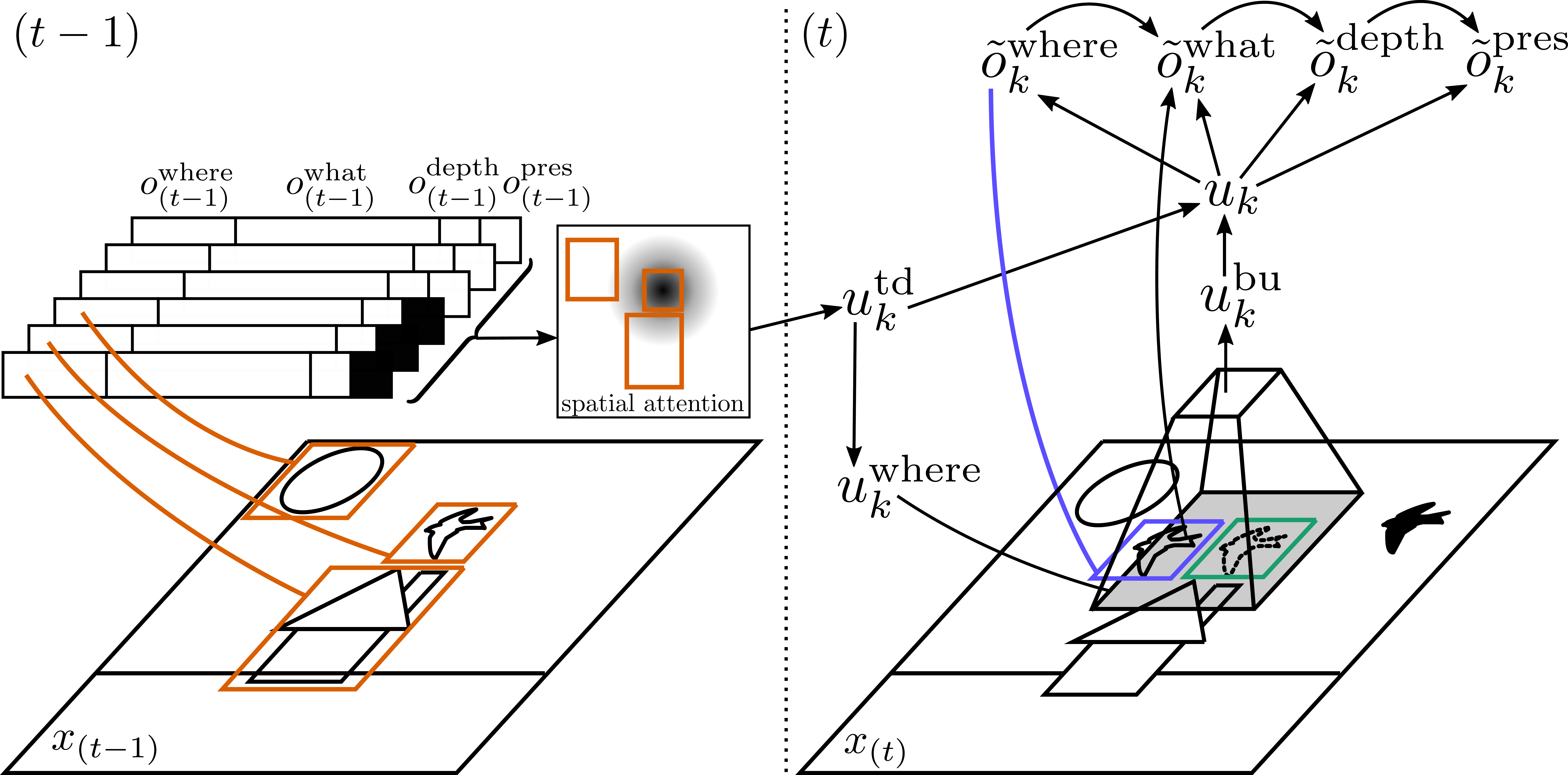}
                \caption{
                    Schematic depicting the propagation module updating an object with index $k$, tracking the location of the white bird. A feature vector for the object, which also takes into account nearby objects, is first created using spatial attention. Next, an initial glimpse (grey region) is specified with respect to the object's location from the previous time step (green box). This glimpse is then processed by a neural network (trapezoid) and used to predict and apply an update to the $\textit{where}$ attribute, resulting in $\sni{\tilde{o}}{where}{k}$. The blue box corresponds to the location of the image referred to by $\sni{\tilde{o}}{where}{k}$. Another glimpse is extracted at location $\sni{\tilde{o}}{where}{k}$, and updates to the remaining attributes are predicted and applied autoregressively. Here we have omitted the latent propagation variables $\tilde{z}_{k}$. \label{fig:propagation}
                }
            \end{figure}

            Propagation at time $t$ takes in the current frame $x_{(t)}$ and the objects from the previous timestep $o_{(t-1)}$, and propagates the objects forward in time, using information from the current frame to update the object attributes.
            Let $K$ be the number of objects from the previous timestep. In this section, all variables are matrices with $K$ as their leading dimension.
            All networks are fully-connected networks that are applied to each matrix row independently (i.e. ``object-wise'') and in parallel. This is similar to the convention used in the discovery module, but with the ``object'' dimension $K$ taking the place of the two spatial dimensions $(H, W)$.
            The structure of propagation for an individual object is shown in Figure \ref{fig:propagation}.
            Here we outline the major points of the propagation module; for additional details, see Section A.1 of the Appendix.
            To reduce visual clutter, assume that all variables have temporal index $t$ unless otherwise specified.

            We begin by computing a feature vector for each object in $o_{(t-1)}$. In order to handle interactions between objects, we can have the feature vector for an object depend on other objects as well. Here we make the assumption that object interactions are spatially local (which is enough to handle collisions, for example). Thus we compute features using a spatial attention step similar to the one used in the Discovery module, allowing the features for an object to depend on attributes of nearby objects:
            \begin{align*}
                \sn{u}{td} = \text{SpatialAttention}^\text{prop}_\phi(o_{(t-1)}, \sigma)
            \end{align*}
            Next we need to condition attribute updates on the current frame. Rather than condition on the entire frame, we instead extract a set of glimpses in a region near each object's location from the previous timestep:
            \begin{align*}
                \sn{u}{where} &= \sni{o}{where}{(t-1)} + 0.1 \cdot \sni{p}{glimpse}{\phi}(\sn{u}{td})\\
                \tilde{g} &= \tau(x, \sn{u}{where})\\
                \sn{u}{bu} &= \sni{p}{bu}{\phi}(\tilde{g})
            \end{align*}
            A network $\sni{p}{fuse}{\phi}$ then combines bottom-up and top-down information:
            \begin{align*}
                u = \sni{p}{fuse}{\phi}(\sn{u}{bu}, \sn{u}{td})
            \end{align*}
            From here we autoregressively predict new values for the object attributes; this is similar to attribute prediction in the discovery module, except that rather than directly predicting attribute values, we predict attribute \textit{updates} and subsequently apply them:
            \begin{align*}
                \sn{\tilde{\mu}}{where}, \sn{\tilde{\sigma}}{where} &= \sni{p}{where}{\phi}(u)\\
                \sn{\tilde{z}}{where} &\sim N(\sn{\tilde{\mu}}{where}, \sn{\tilde{\mu}}{where})\\
                \sn{\tilde{o}}{where} &= f^\text{where}(\sni{o}{where}{(t-1)}, \sn{\tilde{z}}{where})
            \end{align*}
            Next we extract and process another set of glimpses at $\sn{\tilde{o}}{where}$:
            \begin{align*}
                \tilde{g}' &= \tau(x, \sn{\tilde{o}}{where})\\
                \sn{u}{obj} &= \sni{p}{obj}{\phi}(\tilde{g}')
            \end{align*}
            Finally we predict updates to the remaining attributes:
            \begin{align*}
                \sn{\tilde{\mu}}{what}, \sn{\tilde{\sigma}}{what} &= \sni{p}{what}{\phi}(u, \sn{u}{obj}, \sn{\tilde{o}}{where})&\\
                \sn{\tilde{z}}{what} &\sim N(\sn{\tilde{\mu}}{what}, \sn{\tilde{\sigma}}{what})&\\
                \sn{\tilde{o}}{what} &= f^\text{what}(\sni{o}{what}{(t-1)}, \sn{\tilde{z}}{what})\\
                \sn{\tilde{\mu}}{depth}, \sn{\tilde{\sigma}}{depth} &= \sni{p}{depth}{\phi}(u, \sn{u}{obj}, \sn{\tilde{o}}{where}, \sn{\tilde{o}}{what})\\
                \sn{\tilde{z}}{depth} &\sim N(\sn{\tilde{\mu}}{depth}, \sn{\tilde{\sigma}}{depth})\\
                \sn{\tilde{o}}{depth} &= f^\text{depth}(\sni{o}{depth}{(t-1)}, \sn{\tilde{z}}{depth})\\
                \sn{\tilde{\mu}}{pres} &= \sni{p}{pres}{\phi}(u, \sn{u}{obj}, \sn{\tilde{o}}{where}, \sn{\tilde{o}}{what}, \sn{\tilde{o}}{depth})\\
                \sn{\tilde{z}}{pres} &\sim \text{Logistic}(\sn{\tilde{\mu}}{pres})\\
                \sn{\tilde{o}}{pres} &= \sni{o}{pres}{(t-1)} \cdot \text{sigmoid}(\sn{\tilde{z}}{pres})
            \end{align*}
            Notice that propagation cannot increase the value of the \textit{pres} attribute, due to the form of the update (multiplication by a sigmoid). This ensures that objects are only ever discovered by the Discovery module, which is better equipped for it.
            $f^\text{where}$, $f^\text{what}$ and $f^\text{depth}$ determine the forms of the other updates; full details are left to the Appendix. 

            Propagation in SILOT is similar to propagation in SQAIR, with one significant exception. In SQAIR, objects within a timestep are updated sequentially; this allows objects within a timestep to condition on one another, facilitating coordination between objects and supporting behavior such as explaining away. However, this sequential processing can be computationally demanding when there are large numbers of objects. In contrast, SILOT updates all objects within a timestep in parallel; a degree of coordination between objects is achieved via the spatial attention step.

    \subsection{Selection}\label{sec:select}
        Past architectures in this domain, particularly AIR and SQAIR, use discrete Bernoulli random variables for their equivalents of the \textit{pres} attribute, and are thus forced to employ reinforcement learning--inspired techniques for estimating gradients. In order to avoid this complication and make our architecture fully differentiable, we opted to model $\pres{o}$ as real-valued BinConcrete random variables, which are differentiable relaxations of Bernoullis \citep{maddison2016concrete}. One downside of this choice is that objects never ``go away'', even if $\pres{o} = 0$; all posited objects are always present, but to varying degrees. If care is not taken this will result in scaling issues, since the discovery module yields $HW$ new objects per frame; if all these objects are kept and propagated forward, we would end up with a collection of $HW(t+1)$ objects at timestep $t$, which would quickly become intractable.
        
        To fix this, we use a simple top-$K$ selection strategy wherein we keep only the $K$ objects from the union of $\tilde{o}_{(t)}$ and $\bar{o}_{(t)}$ with highest values for the $\textit{pres}$ attribute, for fixed integer $K$. While this hard selection step is not differentiable, the number of objects with non-negligible values for the \textit{pres} attribute is intended to be small compared to $HW(t+1)$ (see Section \ref{sec:prior}), and we have not found this non-differentiability to cause problems in training as long as $K$ is large enough. As a rule-of-thumb, we typically set $K$ to be roughly 25\% larger than the maximum number of objects that we expect to see in a single frame; this ensures there is always room for a reasonable number of objects with low values for \textit{pres} to be propagated and receive gradient feedback. The output of this selection step is the final set of objects $o_{(t)}$.

    \subsection{Rendering}\label{sec:render}
        The differentiable rendering module is the sole constituent of the VAE decoder $g_\theta(x | z)$, taking in the current set of objects $o_{(t)}$ and yielding a reconstructed frame $\hat{x}_{(t)}$. The rendering process is highly structured, and this structure gives meaning to the object attributes. Object appearance and transparancy are predicted from $\sn{o}{what}$, the object is placed at location $\sn{o}{where}$ via spatial transformers, $\sn{o}{depth}$ parameterizes a differentiable approximation of relative depth, and object transparency is multiplied by $\sn{o}{pres}$ so that objects are only rendered to the extent that they exist. Full details can be found in Section A.2 of the Appendix.

    \subsection{Training}
        Recall that propagation, discovery and selection form the VAE encoder $h_\phi(z | x)$, while rendering forms the VAE decoder $g_\theta(x | z)$. The network is trained by maximizing the evidence lower bound \citep{kingma2013auto}:
        \begin{gather}
            \mathcal{L}(\phi, \theta) \coloneqq E_{x \sim h(x), z \sim h_\phi(z|x)} \left[ \log\left(\frac{g_\theta(x | z)g(z)}{h_\phi(z | x)}\right) \right]\label{eq:train}\\
            \phi^*, \theta^* = \argmax_{\phi, \theta} \mathcal{L}(\phi, \theta)
        \end{gather}
        Here $h(x)$ is the distribution over videos defined by the dataset, and $g(z)$ is the prior distribution over $z$. The optimization is performed by gradient ascent.

        \noindent \textbf{Curriculum Learning.} Following SQAIR, we train the network with a form of curriculum learning. We begin by training on only the first 2 frames of each video. We then increase the number of training frames by 2 every $N_{\text{curric}}$ update steps. After $(\lceil T / 2 \rceil - 1)N_{\text{curric}}$ update steps the network will be training on complete videos. 
        This was observed to help with training stability, possibly because it allows the network to learn object appearances before learning object dynamics.

        \noindent \textbf{Discovery Dropout.} Early in development we found the network often tried to predict objects for new frames exclusively by way of the discovery module rather than propagating objects from the previous frame. This strategy can yield reasonable reconstruction performance, but will fail at object tracking since object identities are not maintained from one frame to the next. Moreover, this is a local minimum; once the network starts relying completely on discovery, propagation stops being trained and cannot improve. To discourage this behavior, we designed a technique that we call \textit{discovery dropout}. Each timestep other than $t = 0$, the entire discovery module is turned off with probability $p_\text{dd}$. This forces the network do as much as possible through propagation rather than discovery, since the network is never sure whether the discovery module will be turned on for the next timestep. Throughout this work we use $p_\text{dd} = 0.5$.
    \begin{figure*}[ht]
        \centering
        \includegraphics[width=0.8\textwidth]{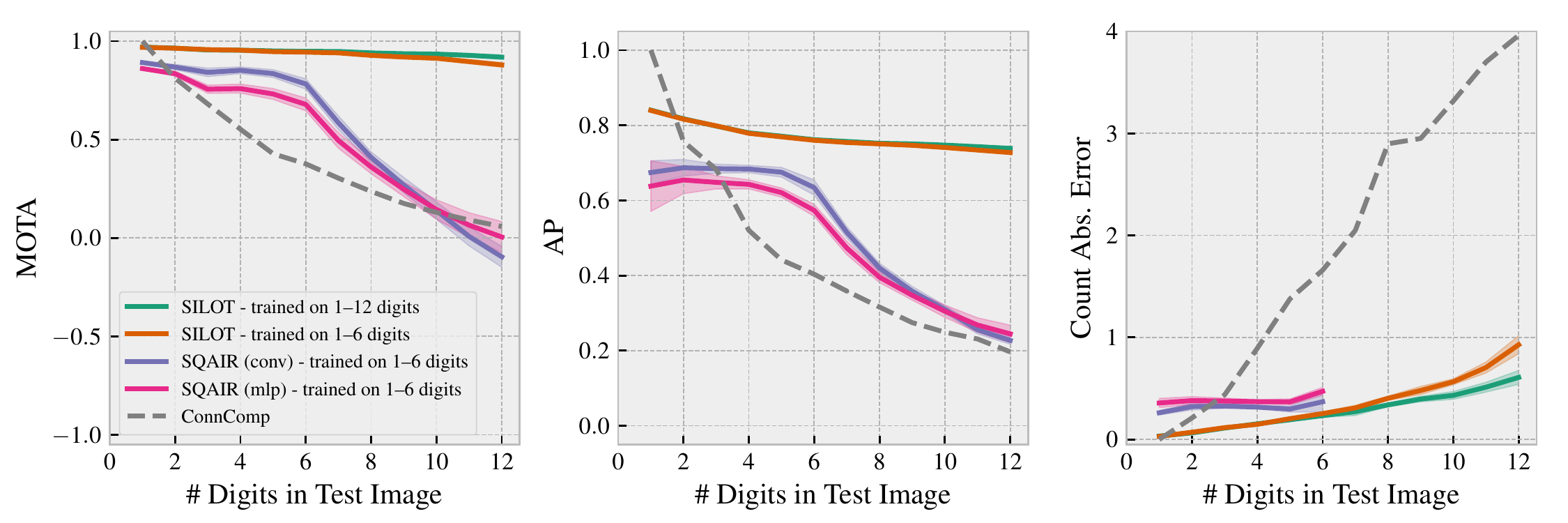}
        \caption{Probing object tracking performance as number of digits per video varies in the Scattered MNIST task. All points are averages over 6 random seeds, filled regions are standard deviations (except for the fully deterministic ConnComp algorithm).\label{fig:mnist_plot}
        }
    \end{figure*}

    \subsection{Prior Distribution}\label{sec:prior}
       A crucial component of a VAE is the prior over latent variables, $g(z)$, which can be used to influence the statistics of latent variables yielded by the encoder. For the majority of the latents, we assume independent Normal distributions. However, for the Logistic random variables $\sn{\bar{z}}{pres}$ and $\sn{\tilde{z}}{pres}$ we design a prior that puts pressure on the network to reconstruct the video using as few objects as possible (i.e. few objects with large values for $\sn{o}{pres}$). This pressure is necessary for the network to extract quality object-like representations; without it, the network is free to set all $\sn{o}{pres} = 1$, and the extracted objects become meaningless. The prior is identical to the one used in SPAIR \citep{crawford2019spatially}, applied to the union of the discovered and propagated objects for a given timestep.

       As done in SQAIR, we can also use a learned prior $g_\theta(z)$, whose parameters can be trained alongside the other parameters. In the current work, we do this only for the propagation module. In effect, this amounts to learning a separate prior propagation module which does not take into account information from the current frame. This prior propagation module employs an architecture that is similar to the main propagation module, except that it omits the glimpse components which extract information from the current frame. The prior propagation module thus is forced to truly learn the dynamics of the objects, whereas the main propagation module can rely on information from the current frame. In the training objective (Equation \eqref{eq:train}), we use the static prior $g(z)$ for the discovery latents, and an even mixture of the static and learned priors for the propagation latents.

\section{Experiments}
    We tested SILOT in a number of challenging object discovery and tracking tasks, emphasizing large videos containing many objects. We use 3 metrics to assess model performance: Multi-Object Tracking Accuracy (MOTA) (with IoU threshold 0.5), a standard measure of object tracking performance \citep{milan2016mot16}; Average Precision (AP) (with IoU=0.1:0.1:0.9), a standard measure of object detection performance \citep{everingham2010pascal}; and Count Abs. Error, or the absolute difference between the number of objects predicted by the model and the ground truth number of objects (on a frame-by-frame basis). Intersection over Union (IoU) is a measure of the overlap between two objects.

    We compare against a simple baseline algorithm called ConnComp (Connected Components), which works by treating each connected region of similarly-colored pixels in a frame as a separate object, and computing object identities over time using the Hungarian algorithm \citep{kuhn1955hungarian}. The performance of ConnComp can be interpreted as a measure of the difficulty of the dataset; it will be successful only to the extent that objects can be tracked by color alone.

    Code for running these experiments is available online at \mbox{\url{https://github.com/e2crawfo/silot}}.
    To supplement the quantitative results provided in this section, we also provide qualitative results:
    videos depicting the performance of trained SILOT networks can be found at \mbox{\url{https://sites.google.com/view/silot}}, and static visualizations are shown in the Appendix.

    \begin{figure*}[ht]
        \centering
        \includegraphics[width=0.8\textwidth]{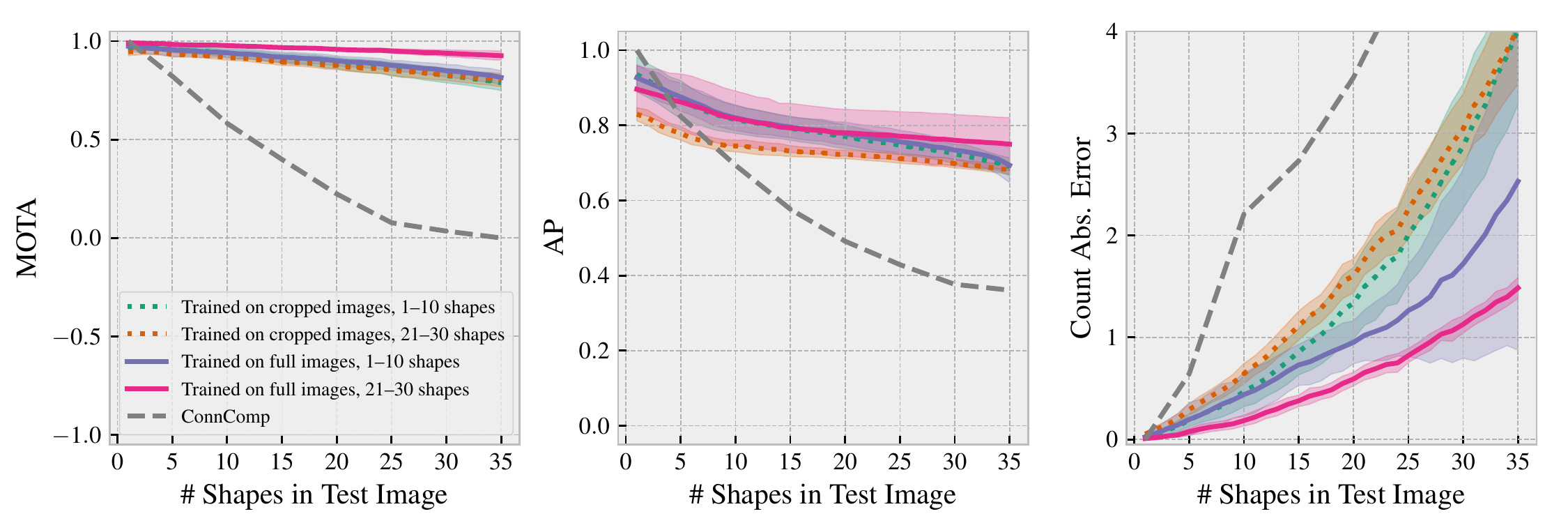}
        \caption{Probing SILOT's object tracking performance as the number of shapes per video varies in the Scattered Shapes task. All points are averages over 4 random seeds, and filled regions are standard deviations (except for the fully deterministic ConnComp algorithm). We manipulated 2 different aspects of the training setup: training on images that contain 1--10 shapes vs 21--30 shapes, and training on random $60 \times 60$ crops vs. full $96 \times 96$ images (full videos are always used at test time).\label{fig:shapes_plot}
        }
    \end{figure*}

    \subsection{Scattered MNIST}
        In this experiment, each 8-frame video is generated by first selecting a number of MNIST digits to include and sampling that number of digits. Videos have spatial size $48 \times 48$ pixels, digits are scaled down to $14 \times 14$ pixels, and initial digit positions are randomly chosen so that the maximum overlap allowed is small but non-trivial. For each digit we also uniformly sample an initial velocity vector with a fixed magnitude of 2 pixels per frame. Digits pass through one another without event, and bounce off the edges of the frames.
        
        In order to test generalization ability, we used two training conditions: training on videos containing 1--6 digits, and training on videos containing 1--12 digits. In both cases, we test performance on videos containing up to 12 digits; networks trained in the 1--6 condition will thus be required to generalize beyond their training experience.

        We compare SILOT against SQAIR.
        We tested SQAIR with two different discovery backbone networks (the network which first processes each input frame): an MLP, and a convolutional network with the same structure as SILOT's convolutional backbone $d^\text{bu}_\phi$.
        Note that using a convolutional network as a backbone, rather than an MLP, should improve SQAIR's spatial invariance to a degree; however it will still lack, among other features, SILOT's local object specification scheme, which is part of what allows SILOT's discovery module to be interpreted as an array of local discovery units.
        
        We also experimented with Tracking by Animation (TbA) \citep{he2018tracking}, but were unable to obtain good tracking performance on these densely cluttered videos. One relevant point is that TbA lacks a means of encouraging the network to explain scenes using few objects, and we found TbA often using several internal objects to explain a single object in the video;
        in contrast, both SILOT and SQAIR use priors on $\sn{o}{pres}$ which encourage $\sn{o}{pres}$ to be near 0, forcing the networks to use internal objects efficiently.

        Results are shown in Figure \ref{fig:mnist_plot}. We have omitted both SQAIR networks trained on the 1--12 digit condition; SQAIR's simpler discovery module was unable to handle the more densely packed scenes, resulting in highly varied performance that would make the plot unreadable. From the plot we can see that SILOT outperforms SQAIR by a large margin (especially when extrapolating beyond training experience), and that SILOT trained on 1--6 digits suffers only a minor performance hit compared to SILOT trained on 1--12 digits, a result of SILOT's spatially invariant architecture. In the Appendix, we also show results obtained when using the learned prior propagation module (see Section \ref{sec:prior}) in place of the regular propagation module.

        Note the slight improvement in performance of the AP of the SQAIR networks when going from 1 to 2 digits. This reflects an interesting strategy developed by the SQAIR networks. On the first timestep, they often turn ``on'' several more objects than are in the video; this pushes down the AP score, with a greater negative effect when there are fewer objects. We hypothesize that it settles on this strategy because the backbone networks are unable to get an accurate read on the number of objects present, and so the network settles for a strategy of casting a ``wide net'' on the first timestep.

        Additionally, when testing on videos containing 1--6 digits, the SQAIR networks were used in the same fashion as they were during training. However, when testing on videos containing 7 or more digits, we found that the network's performance at guessing the number of objects to use became very poor; thus, for such test cases we artificially told the network the correct number of objects to use by running SQAIR's recurrent network for the correct number of steps.

    \subsection{Scattered Shapes}
        In this experiment we tackle larger images with significantly more objects to track. Here each video has spatial size $96 \times 96$, and contains a number of moving monochrome shapes. We use 6 different colors and 5 different shapes. Initial shape velocities are uniformly sampled with a fixed magnitude of 5 pixels per frame.

        Here we test ability to generalize to scenes that are larger and/or contain different numbers of objects than scenes encountered during training. To assess the former, networks were trained on either random $60 \times 60$ crops of input videos, or full $96 \times 96$ videos (note that SILOT's fully convolutional discovery module allows it to process videos of any size). To assess the latter, networks were trained on videos containing either 1--10 shapes or 21--30 shapes. This yields 4 different training conditions. At test time, we use full $96 \times 96$ videos containing up to 35 shapes.

        Results are shown in Figure \ref{fig:shapes_plot}. There we see that all SILOT networks drastically outperform the baseline. Additionally, there is significant evidence that the networks are able to generalize well to both different numbers of shapes and different sized videos than what was encountered during training. In particular, the network trained on $60 \times 60$ crops of videos containing 1--10 shapes achieves reasonable performance when asked to process $96 \times 96$ images containing up to 35 shapes.

    \subsection{Atari}
        We also applied SILOT to videos obtained from select Atari games \citep{bellemare2013arcade} (the possibility of applying a SQAIR-style model to reinforcement learning was first suggested in \citep{kosiorek2018sequential}). Results are shown in Table \ref{table:atari}. Ground truth object labels for evaluation were obtained using the ConnComp algorithm, which is reasonably effective for the chosen games.
        Our goal here was to push scalability, rather than ability to deal with e.g. overlapping objects.
        \begin{table}
            \centering
            \begin{tabular}{c|c|c|c}
                Game & MOTA & AP & Count Abs. Error\\
                \hline
                \hline
                Space Invaders & .89 & .73  & 2.94 \\
                Asteroids & .67 & .67 & 1.81 
            \end{tabular}
            \caption{
                SILOT performance on Atari videos.
                \label{table:atari}
            }
       \end{table}

\section{Conclusion}
    In the current work we proposed an architecture for scalable unsupervised object tracking, and empirically demonstrated a number of its benefits. Many interesting directions are left for future research. For one, it should be possible to extend the discovery module to discover objects at different scales, similar to the supervised object detection architecture SSD \cite{liu2016ssd}. Additionally, in order to have SILOT and related approaches work on real videos, it will be necessary to include a means of dealing with complex, dynamic backgrounds; past work in this area has generally preprocessed videos with an off-the-shelf background subtractor, largely avoiding the problem of modeling the background (though see concurrent work in \cite{jiang2019scalable} which makes significant progress on this issue). Finally, as suggested in \citep{kosiorek2018sequential}, these architectures are quite complex, and it would be worthwhile to perform ablation studies to determine whether all components are necessary for good performance. 

\setcounter{secnumdepth}{0}

\section{Acknowledgements}
Funding for this work was provided by Samsung.

{
\fontsize{9.0pt}{10.0pt} \selectfont
\bibliography{lib}
\bibliographystyle{aaai}
}

\setcounter{secnumdepth}{2}

\appendix
\renewcommand{\theequation}{\thesection\arabic{equation}}
\renewcommand\thefigure{\thesection\arabic{figure}}

\section{Module Details}
    \subsection{Propagation}
        The majority of the propagation module was described in Section 3.4. Here we provide a few additional details.

        \noindent \textbf{Attribute Updates.}
        In Section 3.4 we described functions $f^\text{where}$, $f^\text{what}$ and $f^\text{depth}$ which perform propagation updates:
        \begin{align*}
            f^{y/x}(o_{(t-1)}^{y/x}, \tilde{z}^{y/x}) &= o_{(t-1)}^{y/x} + \text{tanh}(\tilde{z}^{y/x})\\
            f^{h/w}(o_{(t-1)}^{h/w}, \tilde{z}^{h/w}) &= a^{h/w} \cdot \text{sigm}(\text{sigm}^{-1}(o_{(t-1)}^{h/w}) + \tilde{z}^{h/w})\\
            f^\text{depth}(o_{(t-1)}^\text{depth}, \tilde{z}^\text{depth}) &= \text{sigm}(\text{sigm}^{-1}(o_{(t-1)}^\text{depth}) + \tilde{z}^\text{depth})
        \end{align*}
        Here $\text{sigm}$ stands for the sigmoid function. Finally, $f^\text{where}$ is a complicated function (imported from SQAIR) that is difficult to summarize succinctly here; see the accompanying code for details.
        Investigating whether the \textit{where} update can be replaced with something simpler should be a target of future ablation studies.

        \noindent \textbf{Recurrent hidden state.}
        For propagation we assume an additional object attribute $\sni{o}{hidden}{(t)}$, which stores the hidden state of a recurrent neural network $\sni{p}{rnn}{\phi}$ and is fully deterministic.
        This hidden state of each object is provided as an additional argument to the spatial attention module which builds features for propagation.
        After each propagation step, the hidden state for each object is updated independently by running the recurrent network, taking the new object attributes as input:
        \begin{align*}
            \sn{\tilde{o}}{hidden} &= \sni{p}{rnn}{\phi}(\sni{o}{hidden}{(t-1)}, \sn{\tilde{o}}{where}, \sn{\tilde{o}}{what}, \sn{\tilde{o}}{depth}, \sn{\tilde{o}}{pres})
        \end{align*}
        We can think of the hidden state as providing the propagation module with a deterministic path from an object's past to the present. Newly discovered objects are given a default initial hidden state. The value of the default hidden state is a trainable parameter.

    \subsection{Rendering}
        The rendering module is the sole constituent of the VAE decoder. It takes in the current set of objects $o_{(t)}$ and renders them into a frame. We start by focusing on a single object with index $k$, and drop temporal indices. First, an appearance map and a partial transparancy map are predicted: 
        \begin{align*}
            \beta^\text{logit}_k, \xi^\text{logit}_k = \sni{r}{obj}{\theta}(\sn{o}{what}_k)\\
            \beta_k = \text{sigmoid}(\mu^\beta + \sigma^\beta \beta^\text{logit}_k)\\
            \xi_k = \text{sigmoid}(\mu^\xi + \sigma^\xi \xi^\text{logit}_k)
        \end{align*}
        The appearance map $\beta_k$ has shape $(H_\text{obj}, W_\text{obj}, 3)$, while the partial transparency map $\xi_k$ has shape $(H_\text{obj}, W_\text{obj}, 1)$, for integers $H_\text{obj}$, $W_\text{obj}$.
        Meanwhile $\sigma^\beta, \mu^\beta, \sigma^\xi \text{ and } \mu^\xi$ are scalar hyperparameters that can be used to control the relative speed with which appearance and transparency are trained.

        $\xi_k$ is multiplied by $\sn{o}{pres}_k$ to ensure that objects are only rendered to the image to the extent that they are present, yielding a final transparency map:
        \begin{align*}
            \alpha_k = \xi_k \cdot \sn{o}{pres}_k
        \end{align*}
        Next we combine $\alpha_k$ with $\sn{o}{depth}_k$ to get an \textit{importance} map:
        \begin{align*}
            \gamma_k = \alpha_k \cdot \sn{o}{depth}_k
        \end{align*}
        For each object, an inverse spatial transformer parameterized by $\sn{o}{where}_k$ is then used to create image-sized versions of these three maps, with the input maps placed in the correct location:
        \begin{align*}
            \alpha'_k, \beta'_k, \gamma'_k = \tau^{-1}([\alpha_k, \beta_k, \gamma_k], \sn{o}{where}_k)
        \end{align*}
        For each location in one of these image-sized maps, the value is obtained from a corresponding location in the input space, dictated by the location parameters $\sn{o}{where}_k$. However, some of these locations will lie ``outside'' of the input map, and for these locations we use a default value. In particular, we use a default of 0 for $\alpha$ and $\beta$, and $-\infty$ for $\gamma$.

        To obtain the output frame, the image-sized appearance maps are combined by weighted summation. For each pixel we take the softmax (over objects) of the importance values, and weight each object by the resulting value. We also weight by $\alpha'$ to implement transparency. Thus we have:
        \begin{align*}
            \hat{x}_{(t)} = \frac{\sum_{k=0}^{K-1} \beta'_k \alpha'_k e^{\gamma'_k / \lambda}}{\sum_{\ell=0}^{K-1} e^{\gamma'_{\ell} / \lambda}}
        \end{align*}
        Here $\lambda$ is a hyperparameter that acts as the temperature of the softmax. When a given pixel is within the bounding boxes of two or more objects, the softmax implements a differentiable approximation of relative depth, and objects with larger values for $\sn{o}{depth}$ (and thus larger values for $\gamma$, all else being equal) are rendered on top of objects with lower values. The default of $-\infty$ used when spatially transforming $\gamma$ ensures that objects do not contribute to the softmax for pixels that are not within their bounding box.

        Note that all computations in this section can be parallelized to a high degree (i.e. across objects).
        However, for large frames this scheme can still be expensive in terms of both memory and computation.
        Thus in practice we use an equivalent (but more complex) implementation that avoids explicitly constructing image-sized maps for each object.
        The output of rendering is an image with dimensions $(H_\text{inp}, W_\text{inp}, 3)$; to obtain $g_\phi(x | z)$, we use this image to parameterize a set of (conditionally) independent Bernoulli random variables, one for each pixel and channel.

\section{Spatial Attention}
    In this section we provide details on the spatial attention steps used in the discovery and propagation modules. Both are similar to Neural Physics Engine \citep{chang2016compositional}, though more so for the Propagation version.

    \subsection{Spatial Attention for Discovery}
        Recall that in the discovery module, spatial attention is used to obtain, for each discovery unit, a feature vector summarizing nearby propagated objects. The main motivation is to allow the discovery module to avoid rediscovering objects that are already accounted for:
        \begin{align*}
            \sni{v}{td}{(t)} &= \text{SpatialAttention}^\text{disc}_\phi(\tilde{o}_{(t)}, \sigma)
        \end{align*}
        We first narrow our focus to a single discovery unit with indices $ij$. For every propagated object $\tilde{o}_k$ (dropping temporal indices here), we first use an MLP $\sni{d}{spatial}{\phi}$ to compute a feature vector that is specific to $ij$. As input to this MLP, we supply the object $\tilde{o}_k$, except that we replace the position attributes $\tilde{o}^y_k$, $\tilde{o}^x_k$ with \textit{relative} position attributes $\tilde{o}^{y'}_{ij,k}$ and $\tilde{o}^{x'}_{ij,k}$. Here we are assuming that what is important is the position of the propagated objects relative to the grid cell, rather than their absolute positions. Noting that the center of grid cell $ij$ has location
        \begin{align*}
            ((i+0.5) \cdot c_h, (j + 0.5) \cdot c_w)
        \end{align*}
        we have
        \begin{align*}
           \tilde{o}^{y'}_{ij,k} = \tilde{o}^y_k - (i + 0.5) \cdot c_h\\
           \tilde{o}^{x'}_{ij,k} = \tilde{o}^x_k - (j + 0.5) \cdot c_w
        \end{align*}
        The spatial attention module then just sums these feature vectors over propagated objects $k$, weighted by a Gaussian kernel:
        \begin{align*}
            \sni{v}{td}{ij} &= \sum_{k=0}^{K-1} G(\tilde{o}^{y'}_{ij,k}, \tilde{o}^{x'}_{ij,k}, \sigma) \cdot \sni{d}{spatial}{\phi}(\tilde{o}^{\setminus yx}_k, \tilde{o}^{y'}_{ij,k}, \tilde{o}^{x'}_{ij,k})
        \end{align*}
        where $G$ is the density of a 2 dimensional Gaussian, and $\tilde{o}^{\setminus yx}_k$ contains all attributes of $\tilde{o}_k$ except $y$ and $x$. Note that this can be computed for all $ij$ and $k$ in parallel (except for the summation over $k$).

    \subsection{Spatial Attention for Propagation}
        In the propagation module, spatial attention is used to compute a feature vector for each object from the previous step, which is subsequently used to predict updates to the attributes of the object.
        \begin{align*}
            \sni{u}{td}{(t)} = \text{SpatialAttention}^\text{prop}_\phi(o_{(t-1)}, \sigma)
        \end{align*}
        This vector is supposed to take into account attributes of the object itself, as well as attributes of nearby objects. This should allow the updates to take into account the effect of nearby objects on the target object.

        We first narrow our focus to a target object with index $\ell$. We use an MLP $\sni{p}{td}{\phi}$ to compute an initial feature vector:
        \begin{align*}
            \sni{u}{td$'$}{\ell} = \sni{p}{td}{\phi}(o_\ell)
        \end{align*}
        Then for every object $k$ we compute the position of object $k$ relative to object $\ell$:
        \begin{align*}
           o^{y'}_{\ell,k} = o^y_k - o^y_\ell\\
           o^{x'}_{\ell,k} = o^x_k - o^x_\ell
        \end{align*}
        Next we use an MLP $\sni{p}{spatial}{\phi}$ to get feature vectors for $o_k$ in the context of target object $o_\ell$. This is similar to discovery, except the MLP also takes $\sni{u}{td$'$}{\ell}$ as an argument. The results are summed and weighted by a Gaussian kernel, and then $\sni{u}{td$'$}{\ell}$ is added in:
        \begin{align*}
            \sni{u}{td}{\ell} &= \sni{u}{td$'$}{\ell} + \sum_{k=0}^{K-1} G(o^{y'}_{\ell,k}, o^{x'}_{\ell,k}, \sigma) \cdot \sni{p}{spatial}{\phi}(o^{\setminus yx}_k, o^{y'}_{\ell,k}, o^{x'}_{\ell,k}, \sni{u}{td$'$}{\ell})
        \end{align*}
        We can think of the second term as computing the additive effect of nearby objects on the target object. Again this can be computed for all $\ell$ and $k$ in parallel (except the summation).

\section{Baseline Algorithm: ConnComp}
    We compare against a simple baseline algorithm called ConnComp (Connected Components), which works as follows. For each frame a graph is created wherein the pixels are nodes, and two pixels are connected by an edge if and only if they are adjacent and have the same color. We then extract connected components from this graph, and call each connected component an object. This yields a set of objects for each frame. In order to track objects over time (i.e. to assign persistent identifiers to the objects), we employ the Hungarian algorithm to find matches between detected objects in each pair of successive frames \citep{kuhn1955hungarian}. As matching cost we use the distance between object centroids, and require that matching objects have the same color (objects pairs with mismatched colors are assigned a cost of $\infty$). The performance of ConnComp can be interpreted as a measure of the difficulty of the dataset; it will be successful only to the extent that objects can be tracked by color alone.

\section{Experiment Details}
    \subsection{Scattered MNIST}
        Each video had spatial size $48 \times 48$ pixels. MNIST digits were resized to $14 \times 14$ pixels (their original size being $28 \times 28$. Initial digit velocities vectors were sampled uniformly, with a fixed magnitude of 2 pixels per frame. Digits passed through one another without event, and bounce of the edges of the frame. Initial digit positions were sampled randomly, one digit at a time. If the cumulative overlap between a newly sampled digit position and existing digit positions exceeded a threshold (98 pixels), the digit position was resampled until the threshold was not exceeded. This allowed us to control the amount of overlap between digits on the first frame. On subsequent frames, no overlap limit is enforced, and indeed because objects pass through one another and bounce of frame borders, the degree of overlap can be quite high in subsequent frames. In such cases, networks need to use information about the locations and trajectories of the overlapping objects at previous timesteps in order to track well.

        The 60,000 MNIST digits were divided up into 80\% training set, 10\% validation set, and 10\% test set. Videos for training set were created by uniformly sampling the number of digits to put in the video (from distribution $\text{Uniform}(1, 6)$ or $\text{Uniform}(1, 12)$ depending on the training condition), and then sampling that number of digits from the set of training digits. Similar approaches were taken for validation and testing; this ensures that at test time, networks are seeing digits that they have never seen before.
        We created 60,000 training videos and 1000 validation videos for each training condition, and 1000 videos for each testing condition (i.e. for each position on the x-axis of the plots in Figure 5 of the main text).

    \subsection{Scattered Shapes}
        Videos in the Scattered Shapes dataset were created in a manner similar to Scattered MNIST, with a few exceptions. Possible colors were red, green, blue, cyan, yellow, magenta, and possible shapes were circle, diamond, star, cross, x. All shape/color combinations were present in training, validation and test datasets. Initial pixel velocities had a fixed magnitude of 5 pixels per frame rather than 2. Additionally, when adding a shape to a video, it's size was randomized by choosing each spatial dimension from a $\text{Normal}(\mu=14, \sigma=1.4)$ distribution.

\section{Model and Training Details}
    \subsection{SILOT}\label{sec:silot-hyper}
        SILOT was implemented in tensorflow, and is available online at \url{https://github.com/e2crawfo/silot}.
        Hyperparameters are listed in Table \ref{table:silot-hyper}.
        Recall that we trained SILOT using a curriculum, starting training on the initial 2 frames of each video, and increasing by 2 frames every $N_\text{curric}$ timesteps until the network is training on full videos. Once the network is training on full frames, we begin an early stopping regime wherein we train until the measure of performance (in most cases, MOTA on the validation set) does not improve for 30,000 training steps (i.e. using a ``patience'' value of 30,000, at the level of training steps rather than epochs), triggering an early stop. Each time an early stop is triggered, we go back to the set of weights that has the best performance so far, divide the learning rate by 3, and resume training. Training ends once early stopping has been triggered 3 times.  We take as our final hypothesis the set of weights that achieved the best performance on the validation set.

        Finally, for the first 1000 update steps we do not backpropagate gradients through \textit{where}, \textit{depth}, or \textit{pres} attributes. Without this initial period, the component networks responsible for encoding and decoding appearance information (particularly $\sni{d}{obj}{\phi}$, $\sni{p}{obj}{\phi}$ and $\sni{r}{obj}{\theta}$) are initially poor at reconstructing any part of the scene, and the network often realizes that it can reduce the reconstruction loss by simply turning all of the objects off or shrinking them into oblivion. Both of these possibilities are very poor local minima, and need to be avoided. Additionally, we use a prior for the \textit{pres} attributes which starts by encouraging the network to use many objects, but over time is annealed into a prior that encourages using few objects (using a very similar approach to SPAIR \citep{crawford2019spatially}).

        SILOT contains a relatively large number of component networks. Table \ref{table:silot-component-networks} lists these networks, along with descriptions of their role and architecture. The architecture of the backbone convolutional network  $\sni{d}{bu}{\phi}$ is:
\begin{verbatim}
    [Conv(n=128, f=4, s=3, nl=RELU)
     Conv(n=128, f=4, s=2, nl=RELU),
     Conv(n=128, f=4, s=2, nl=RELU),
     Conv(n=128, f=1, s=1, nl=RELU), 
     Conv(n=128, f=1, s=1, nl=RELU),
     Conv(n=128, f=1, s=1, nl=None)]
\end{verbatim}
        where $n$ gives the number of filters, $f$ gives the filter width, $s$ gives the stride and $nl$ gives the non-linearity used. No pooling is used at any point.

        \begin{table*} 
            \begin{tabular}{@{}lll@{}}
                \toprule
                \textbf{Description} & \textbf{Variable} & \textbf{Value} \\ \midrule
                 Initial learning rate & & 0.0001 \\
                 Batch size & & 16 \\
                 Max gradient norm & & 10.0 \\
                 Optimizer && Adam \\
                 Patience & & 30,000\\
                 \# of training steps for each stage of curriculum & $N_\text{curric}$ & 40,000\\
                 Probability of discovery dropout & $p_\text{dd}$ & 0.5 \\
                 \midrule
                 Number of propagated/selected objects & $K$ & 16\\
                 Dimension of \textit{what} attribute & $A$ & 64 \\
                 Anchor box size in pixels & $(a_h, a_w)$ & (48, 48) \\
                 Grid cell size in pixels (determined by structure of $d_\phi^\text{bu}$) & $(c_h, c_w)$ & (12, 12) \\
                 Bounds on distance between object center and grid cell center in disc. & $(b^{min}, b^{max})$ & (-0.5, 1.5) \\
                 Standard dev. of Gaussian kernel for spatial attention & $\sigma$ & 0.1 \\
                 \midrule
                 Prior on $\bar{z}^h, \bar{z}^w$ & & $\text{Normal}(\mu=-2.2, \sigma=0.5)$ \\ 
                 Prior on $\bar{z}^y, \bar{z}^x$ &  & $\text{Normal}(\mu=0, \sigma=1)$ \\ 
                 Prior on $\bar{z}^\text{what}$ &  & $\text{Normal}(\mu=0, \sigma=1)$ \\ 
                 Prior on $\bar{z}^\text{depth}$ &  & $\text{Normal}(\mu=0, \sigma=1)$ \\ 
                 Prior on $\bar{z}^\text{pres}$ & & See Section \ref{sec:silot-hyper}\\ 
                 \midrule
                 Prior on $\tilde{z}^h, \tilde{z}^w$ &  & $\text{Normal}(\mu=0, \sigma=0.3)$ \\ 
                 Prior on $\tilde{z}^y, \tilde{z}^x$ &  & $\text{Normal}(\mu=0, \sigma=0.3)$ \\ 
                 Prior on $\tilde{z}^\text{what}$ &  & $\text{Normal}(\mu=0, \sigma=0.4)$ \\ 
                 Prior on $\tilde{z}^\text{depth}$ &  & $\text{Normal}(\mu=0, \sigma=1)$ \\ 
                 Prior on $\tilde{z}^\text{pres}$ & & See Section \ref{sec:silot-hyper}\\ 
                 \midrule
                 Appearance offset and scale & $\mu^\beta, \sigma^\beta$ & (0.0, 2.0) \\
                 Transparency offset and scale & $\mu^\xi, \sigma^\xi$ &  (5.0, 0.1) \\
                 Rendering softmax temperature & $\lambda $ & 0.25 \\
                 Rendered object size & $(H_{obj}, W_{obj})$ & (14, 14) \\
                 \bottomrule
            \end{tabular}
            \caption{
                Base hyperparameter values for SILOT.
                \label{table:silot-hyper}
            }
        \end{table*}

        \begin{table*} 
            \begin{tabular}{@{}lllll@{}}
                \toprule
                 & \textbf{Description} & \textbf{Architecture}\\
                \midrule
                $\sni{d}{bu}{\phi}$ & Computes bottom-up features for disc. grid cells & See Section \ref{sec:silot-hyper}\\
                $\sni{d}{spatial}{\phi}$ & Computes features of propped objects in disc. attention & FC([64, 64], RELU)$^\dagger$ \\
                $\sni{d}{fuse}{\phi}$ & Combines top-down and bottom-up info in disc  & FC([100, 100], RELU)$^\dagger$ \\
                $\sni{d}{where}{\phi}$ & Predicts params. for posterior dist. over $\sn{\bar{z}}{where}$  & FC([100, 100], RELU)$^\dagger$ \\
                $\sni{d}{obj}{\phi}$ & Processes glimpse in disc.  & FC([256, 128], RELU)$^\dagger$ \\
                $\sni{d}{what}{\phi}$ & Predicts params. for posterior dist. over $\sn{\bar{z}}{what}$  & FC([100, 100], RELU)$^\dagger$ \\ 
                $\sni{d}{depth}{\phi}$ & Predicts params. for posterior dist. over $\sn{\bar{z}}{depth}$  & FC([100, 100], RELU)$^\dagger$ \\
                $\sni{d}{pres}{\phi}$ & Predicts params. for posterior dist. over $\sn{\bar{z}}{pres}$  & FC([100, 100], RELU)$^\dagger$ \\
                \midrule
                $\sni{p}{td}{\phi}$ & Computes object features in prop. attention  & FC([64, 64], RELU)\\
                $\sni{p}{spatial}{\phi}$ & Computes object-pair features in prop. attention  & FC([64, 64], RELU)\\
                $\sni{p}{glimpse}{\phi}$ & Predicts params for initial prop. glimpse  & FC([100, 100], RELU) \\
                $\sni{p}{bu}{\phi}$ & Processes initial prop. glimpse  & FC([256, 128], RELU) \\
                $\sni{p}{where}{\phi}$ & Predicts params. for posterior dist. over $\sn{\tilde{z}}{where}$  & FC([100, 100], RELU)\\
                $\sni{p}{obj}{\phi}$ & Processes second glimpse in prop.  & FC([256, 128], RELU) \\
                $\sni{p}{what}{\phi}$ & Predicts params. for posterior dist.  over $\sn{\tilde{z}}{what}$  & FC([100, 100], RELU) \\
                $\sni{p}{depth}{\phi}$ & Predicts params. for posterior dist. over $\sn{\tilde{z}}{depth}$  & FC([100, 100], RELU) \\
                $\sni{p}{pres}{\phi}$ & Predicts params. for posterior dist. over $\sn{\tilde{z}}{pres}$  & FC([100, 100], RELU) \\
                $\sni{p}{rnn}{\phi}$ & Updates deterministic hidden state  & GRU(128) \\
                \midrule
                $\sni{r}{obj}{\theta}$ & Predicts object appearances in rendering.  & FC([128, 256], RELU) \\
                \bottomrule
            \end{tabular}
            \caption{
                Component neural networks in SILOT. FC([N, N], RELU) is a sequence of 3 fully-connected layers (2 hidden layers each with N units, one output layer). The RELU non-linearity is applied only at the hidden layers, the output is left unconstrained.
                $^\dagger$Recall that in Section 3.3 (Discovery) we said that we use convolutional networks in Discovery; however, since those networks generally use a stride and filter size of 1, they are equivalent to applying a single fully-connected layer independently to each spatial location. Thus, here we are listing the fully-connected equivalent of the convolutional networks that were actually used.
                \label{table:silot-component-networks}
            }
        \end{table*}

    \subsection{SQAIR}\label{sec:sqair-hyper}
        For our implementation of SQAIR, we used a lightly modified version of the original implementation:\linebreak \url{https://github.com/akosiorek/sqair}.\linebreak
        Initial tuning of hyperparameters was performed by hand (starting from the default values) until a reasonable range of values for the Scattered MNIST dataset was identified. As model-selection criteria, we used the MOTA (a measure of object tracking performance) averaged over 1--6 digits for the 1--6 training case, and averaged of 1--12 digits for the 1--12 training case. We ran an additional grid search over select hyperparameters.

\section{Experiment Visualizations}
\begin{figure*}
            \centering
            \includegraphics[width=\textwidth]{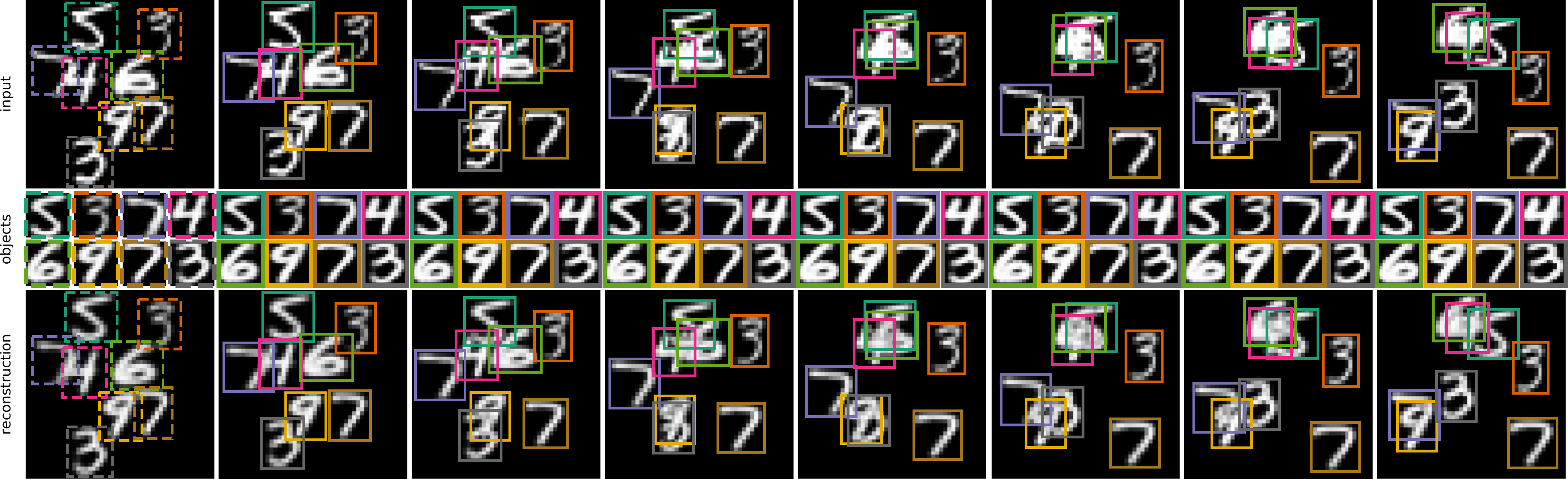}
            \caption{Visualizing a forward pass of a trained SILOT network applied to a video from the Scattered MNIST task containing 8 objects. Top / Bottom: Ground truth / reconstructed frames with bounding boxes for detected objects overlaid. Middle: Predicted appearances for detected objects. Box color represents object identity according to the network. Boxes for objects that SILOT has discovered in a given frame are dashed, while boxes for objects propagated from the previous frame are solid. Notice that the network is able to track objects even after they have passed completely through other objects (e.g. 5 with the green box, 3 with the grey box).\label{fig:mnist_vis}
            }
        \end{figure*}

\begin{figure*}
            \centering
            \includegraphics[width=\textwidth]{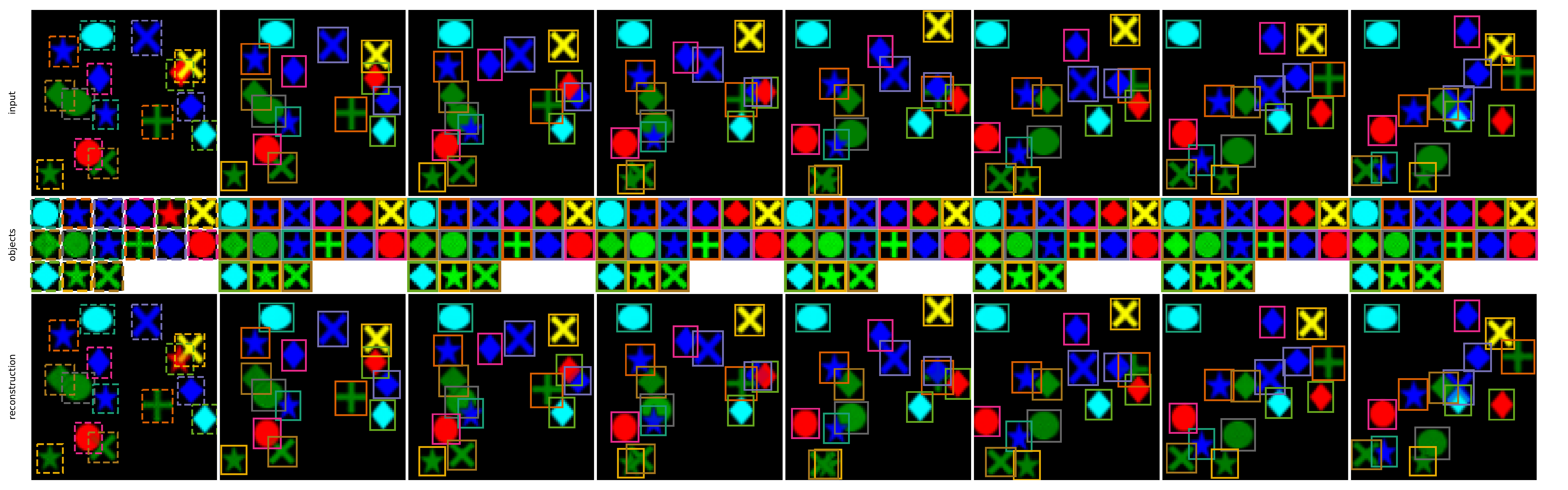}
            \caption{Visualizing a forward pass of a trained SILOT network applied to a video from the Scattered Shapes task containing 15 objects. Top / Bottom: Ground truth / reconstructed frames with bounding boxes for detected objects overlaid. Middle: Predicted appearances for detected objects. Box color represents object identity according to the network. Boxes for objects that SILOT has discovered in a given frame are dashed, while boxes for objects propagated from the previous frame are solid. Notice that the network is able to track objects even after they have been heavily occluded by other objects (e.g. green cross with the orange box that starts near the center).\label{fig:shapes_vis}
            }
        \end{figure*}

        A visualization of a forward pass of SILOT on the Scattered MNIST task is shown in Figure \ref{fig:mnist_vis}, and on the Scattered Shapes task in Figure \ref{fig:shapes_vis}.

\section{Additional Experiments}
    \subsection{Scattered MNIST - Propagating with Prior}
        As detailed in Section 3.8, we trained a learned prior for the propagation latent variables in addition to the static prior. This can be viewed as a duplicate of the main propagation module, except that it does not have access to the frame each time step. Here we test the performance of that module in the Scattered MNIST task. The evaluation procedure is similar to the one used in SQAIR for the same purpose \citep{kosiorek2018sequential}, and runs as follows. For the first 3 timesteps, the regular network is used (with the regular propagation module), in order to discover objects and estimate their initial trajectory. For the remaining 5 frames the discovery module is deactivated, and the prior propagation module is used instead of the main propagation module. We can do this for both SILOT and SQAIR.

\begin{figure}
            \centering
            \includegraphics[width=\columnwidth]{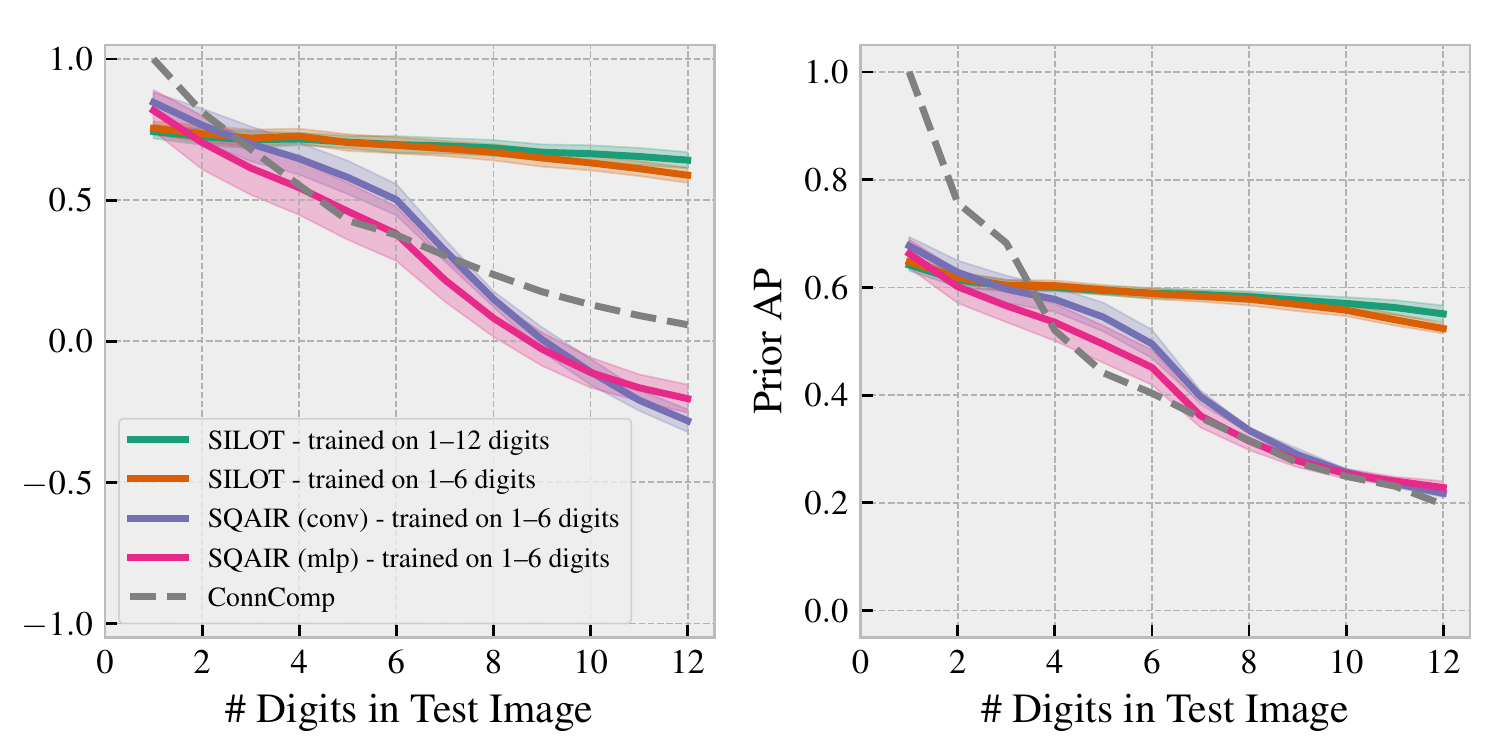}
            \caption{
                Probing ability of the learned prior propagation modules to predict object trajectories without access to the input frames. \label{fig:prior}
            }
        \end{figure}

        Evaluation metrics were computed only on the final 5 frames. We are thus testing the ability of the prior propagation module to predict the trajectories of the objects detected in the first 3 frames by the main network. Results are shown in Figure \ref{fig:prior}.

\end{document}